\documentclass[runningheads,a4paper]{llncs}
\usepackage[utf8]{inputenc}
\usepackage[english]{babel}
\usepackage[T1]{fontenc}
\usepackage{amsfonts}
\usepackage{amsmath}
\usepackage{amssymb}
\usepackage{tikz}
\usepackage{graphicx}
\usepackage{color}
\usepackage{pgfpages}
\usepackage{ulem}
\usepackage{url}

\begin{document}

\title{Compression of Deep Neural Networks\\on the Fly}

\author{Guillaume Soulié, Vincent Gripon, Maëlys Robert}
\institute{Télécom Bretagne}

\date{2015}
\maketitle
\begin{abstract}
Thanks to their state-of-the-art performance, deep neural networks are increasingly used for object recognition. To achieve the best results, they use millions of parameters to be trained. However, when targetting embedded applications the size of these models becomes problematic.
As a consequence, their usage on smartphones or other resource limited devices is prohibited.
In this paper we introduce a novel compression method for deep neural networks that is performed during the learning phase. It consists in adding an extra regularization term to the cost function of fully-connected layers. We combine this method with Product Quantization (PQ) of the trained weights for higher savings in storage consumption. We evaluate our method on two data sets (MNIST and CIFAR10), on which we achieve significantly larger compression rates than state-of-the-art methods.
\end{abstract}

\section{Motivation}

Deep Convolutional Neural Networks (CNNs) \cite{krizhevsky2012imagenet,le1990handwritten,szegedy2014going,simonyan2014very} have become the state-of-the-art for object recognition and image classification. As a matter of fact, most recently proposed systems are using this architecture~\cite{jia2013caffe,donahue2013decaf,simonyan2014very,sermanet2013overfeat,zeiler2013stochastic,gong2014multi}. With this global trend arise questions on how to to import CNNs on embedded platforms~\cite{gokhale2014240}, including smartphones, where data storage and bandwidth are limited. Today the size of a typical CNN is often too large (typically hundred of megabytes for vision applications) for most smartphone users. The purpose of this paper is to propose new techniques for compressing deep neural networks without sacrificing performance.

In this work we focus on compressing CNNs used for vision, although our methodology is not taking any advantage of this particular application field and we expect it to perform similarly on other types of learning tasks.
A typical state-of-the-art CNN \cite{jia2013caffe,sermanet2013overfeat,zeiler2013stochastic} for visual recognition contains several convolutional layers followed by several fully connected layers. For the most challenging datasets, these layers may require hundred of millions of parameters to be trained in order to be efficient. 

These parameters are overparameterized \cite{denil2013predicting} and we aim at compressing them. Note that our motivation is mainly to reduce the model size rather than speeding up the computation time \cite{denton2014exploiting}.

Compressing deep neural networks has been the subject of several recent works.
In \cite{denton2014exploiting} and \cite{jaderberg2014speeding} the authors use compression methods for speeding up CNN testing time.

More recently, some works focus on compressing neural network specially to reduce storage of the network. These works can generally be put into two different categories: some of them focus on compressing the fully connected layers and others on compressing the convolutional layers.
In \cite{gong2014compressing} the authors focus on compressing densely connected layers. In their work, they use signal processing vector quantization methods \cite{jegou2011product,chen2010approximate} such as $k$-means or Product Quantization (PQ).
In \cite{chen2015compressing} the authors focus on compressing the fully connected layers of a Multi-Layer Perceptron (MLP) using Hashing Trick, a low cost hash function to randomly group connection weights into hash buckets, and set the same value to all the parameters in the same bucket. 
In \cite{chen2015compressingConv} the authors propose compressing convolutional layers using a Discrete Cosinus Transform applied on the convolutional filters, followed by Hashing Trick, as for the fully connected layers.

An interesting point showed by \cite{gong2014compressing} is that in a typical sate-of-the-art CNN, more than 90\% of the storage is taken up by the densely connected layers, whereas about 90\% of the running time is taken by the convolutional layers. This is why, in order to compress the size of a CNN, we mainly focus on compressing the densely connected layers.

Instead of using a post-learning method to compress the network, our approach consists in modifying the regularization function used during the learning phase in order to favor quantized weights in some layers -- especially the output ones. To achieve this, we use an idea that was originally proposed in \cite{murray2010algorithm}.
In order to compress furthermore our obtained networks, we also use PQ as described in \cite{gong2014compressing} afterwards. We perform some experiments both on Multi-Layer Perceptrons (MLP) and Convolutionnal Neural Networks. 

In this paper, we introduce a novel strategy to quantize weights in deep learning systems. More precisely:
\begin{itemize}
\item We introduce a regularization term that forces weights to converge to either 0 or 1, before using the product quantization on the trained weights.
\item We show how this extra term impacts performance depending on the depth of the layer it is used onto.
\item We experiment our proposed method on celebrated benchmarks and compare with state-of-the-art techniques.
\end{itemize}
The outline of the paper is as follows. In Section II we discuss related work. Section III introduced our methodology for compressing layers in deep neural networks. In Section IV we run experiments on celebrated databases. Section V is a conclusion.

\section{Related Work}

As already mentioned in the introduction, the densely connnected layers of a state-of-the-art CNN usually involve hundreds of millions of parameters, thus requiring an important storage that may be hard to obtain in practice.
Several works have been published on speeding up CNN prediction speed. In \cite{vanhoucke2011improving} the authors use tricks of CPUs to speed up the execution of CNN. 
In \cite{mathieu2013fast}, the authors show that carrying the convolutionnal operations in the Fourrier domain may lead to a speed-up of 200\%. Two very recent works, \cite{denton2014exploiting,jaderberg2014speeding}, use linear matrix factorization methods for speeding up convolutions and obtain a 200\% speed-up gain with almost no loss in classification accuracy.

The previously mentionned works mainly focus on speeding up the CNN feedforward operations. Recently, several works have been devoted to compressing the CNN size.
In \cite{denil2013predicting}, the authors demonstrate the overparametrization in neural network parameters. Indeed, they show that only 5\% of parameters are enough to accurately predict the 95\% remaining ones.
These results motivate \cite{gong2014multi} to apply vector quantization methods to benefit from redundancy and compress the network parameters.
This compression allows them to obtain results similar to those of \cite{denil2013predicting}: they are able to achieve a compression rate of about 20 without sacrificing accuracy. 
In their paper, they tackle the model size issue by applying PQ on the trained weights.
They are able to achieve a good balance between storage and test accuracy. For the ImageNet challenge ILSVRC2012, they achieve a 16-24 compression rate for the whole network with only 1\% loss on accuracy, using a state-of-the-art CNN.

In \cite{chen2015compressing}, for the first time a learn-based method is proposed to compress neural networks. This method, based on Hashing Trick, allows efficient compression rates. In particular, they show that compressing a large neural network may be more efficient than directly training a smaller one: in their example they are able to divide the loss by two using a eight times larger neural network compressed eight times. The same authors also propose in \cite{chen2015compressingConv} to compress filters in convolutional layers, arguing that the size of the convolutional layers in state-of-the-art's CNN is increasing year after year. Using the fact that learned CNN filters are often smooth, their Discrete Cosinus Transform followed by Hashing Tricks allows them to compress a whole neural network without loosing too much accuracy.

\section{Methodology}
In this section, we present two methods for compressing the parameters in CNN layers. First we introduce the PQ method from~\cite{gong2014compressing}, and then we introduce our proposed learn-based method.

\subsection{Product Quantization (PQ)}
This method has been extensively studied in \cite{gong2014compressing}. The idea is to exploit the inner redundancy of trained weights. In order to do that, the authors propose to use PQ.

PQ consists of partitioning the parameters space into disjoint sub-spaces, and performing quantization in each of them. The term ``product'' refers to the fact that the quantized points in the original parameter space are the cartesian product of the quantized points in each sub-space.
PQ performs increasingly better as the redundancy in each subspace grows.

Specifically, given a layer $L$, let us denote by $W$ the matrix of the corresponding weights and by $(m,n)$ the dimensions of W. Assumming $n$ is divisible by $s$, we can partition $W$ column-wise into $s$ sub-matrices: 

\begin{equation}
W=[W^1,W^2,...,W^s],
\end{equation}
where $W^i \in \mathbb{R}^{m \dot (n/s)}$. 
In \cite{gong2014compressing}, the authors point out that applying PQ on the $x$-axis or the $y$-axis of $W$ does not leads to major diffference in experiments.
We can then perform $k$-means for each sub-matrix $W^i$, i.e. minimize:
\begin{equation}
\sum_{z=1}^{m} \sum_{j=1}^{k} \| w_z^i - c_j^i \|_{2}^{2},
\end{equation}
where $w_i^z$ denotes the $z$-th row of sub-matrix $W^i$, and $c_j^i$ denotes the $j$-th row of sub-codebook
$C^i \in \mathbb{R}^{k \dot (n / s)}$. The $c^i$ which minimize this expression are named centroids. 

Thus, the reconstructed matrix is:
\begin{equation}
\hat{W}=[\hat{W}^1,\hat{W}^2,...,\hat{W}^s],
\end{equation}
where
\begin{equation*}
\hat{w}_j^i = c_j^i, j \text{ being a minimizer of } \min_j \| w_z^i - c_j^i \|_{2}^{2}.
\end{equation*}

We replace $w_j^i$ by $\hat{w}_j^i$ : the nearest centroid of $w_j^i$.
We need to store the nearest centroid indexes for each $w_j^i$ and codebooks of all the $\hat{w}_j^i$ for each sub-vector. The codebook is not negligible, therefore the compression rate is $(32mn)/(\log_2(k)ms + 32kn)$. With a fixed segment size, increasing $k$ will lead to decreasing the compression rate.

\subsection{Proposed method}
Our proposed method is twofold: first, we use a specific added regularization term in order to attract network weights to binary values, then we coarsely quantize the output layers.

Let us recall that training a neural network is generally performed thanks to the minimization of a cost function using a derivative of a gradient descent algorithm. In order to attract network weights to binary values, we add a binarization cost (regularizer) during the learning phase. This added cost pushes weights to binary values. As a result, solutions of the minimization problem are expected to be binary or almost binary, depending on the scaling parameter of the added cost with respect to the initial one. This idea is not new, although we did not find any work applying it to deep learning in the literature. Our choice for the regularization term has been greatly inspired by \cite{murray2010algorithm}.

More precisely, let us denote by $W$ the weights of the neural network, $f(W)$ the cost associated with $W$, $h_W(X)$ and $y(X)$ respectively the output and the label for a given input $X$, we obtain:

\begin{equation}
f(W)=\sum_{X}{\|h_W(X) - y(X)\|_{2}}+ \alpha \sum_{w \in W}{\|w - 1\|_{2}\|w + 1\|_{2}}\;,
\end{equation}

where $\alpha$ is a scaling parameter representing the importance of the binarization cost with respect to the initial cost. Note that possible values for binary weights have been empirically explored and those centered on 0 (here $\{-1,+1\}$) led to the best results. 

Finding a good value for $\alpha$ may be tricky, as a too small value results in a failure of the binarization process and a too large value results in the creation of local minima that will prevent the network from successfully training. To facilitate this selection of $\alpha$, we use a barrier method \cite{murray2010algorithm} that consists in starting with small values of $\alpha$ and incrementing it regularly to help the quantization process. In our experiments, at each iteration, we multiply $\alpha$ by a constant $c = 1.001$.

We observed that some layers are typically very well quantized at the end of this learning phase, whereas others are still far from binary. For that reason we then binarize some of the layers but not all. Again, this selection is made by exploring empirically the possibilities, for example using the results depicted in Figure \ref{pos}.

In order to improve further our compression rate, we then use the PQ method presented in the previous subsection.

The compression rate for our method is $(32mn)/(kn + \log_2(k)ms)$ (instead of $(32mn)/(32kn + \log_2(k)ms)$ for single Product Quantization). With a fixed segment size, increasing $k$ will lead to decreasing the compression rate.

\section{Experiments}
We evaluate these different methods on two image classification datasets : MNIST and CIFAR10. The parameters used for Product Quantizer are a segment size $m$ varying in $\{2,4,5,8\}$ and a number of cluster $k$ varying in $\{4,8,16\}$.

\subsection{Experimental settings}
	\subsubsection{MNIST}
\indent The MNIST database of handwritten digits has a training set of 60,000 examples, and a test set of 10,000 examples. It is a subset of a larger set available from NIST. The digits have been size-normalized and centered in a fixed-size image.
The neural network we use with MNIST is LeNet5. LeNet5 is a convolutional neural network introduced in~\cite{lecun1998gradient}. 


	\subsubsection{CIFAR10}
The CIFAR10 database has a training set of 50,000 examples, and a test set of 10,000 examples. It is a subset of a larger set available from the 80 million tiny images dataset. It consists of 32x32 colour images partitioned into ten classes. 
With the CIFAR10 database, we use a convolutional neural network made of four convolutional layers followed by two fully connected layers. This network has been introduced in \cite{keras}.

\subsection{Layers to quantify}

Our first experiments (with the MNIST database) depicted in Table~\ref{pos} shows the influence of quantified layers on performance. We observe that performance strongly depends on which layers are quantized. More precisely, this experiment shows that one should quantize layers from the output to the input rather than the contrary. This result is not surprising to us as input layers have often been described as similar to wavelet transforms, which are intrinsically analog operators, whereas output layers are often compared to high level patterns which detection in an image is often enough for good classification results.

\begin{figure}[h]
\begin{center}
\begin{tabular}{c|c|c|c||c}
Layer 0 & Layer 1 & Layer 2 & Layer 3 & Test error \\ \hline
- & - & - & - & 0,90\% \\
binarised & - & - & - & 23,19\% \\
binarised & binarised & - & - & 81,26 \% \\
binarised & binarised & binarised & - & 90,26\% \\
binarised & binarised & binarised & binarised & 90,1 \% \\
- & binarised & binarised & binarised & 7,54\% \\
- & - & binarised & binarised & 1,13\% \\
- & - & - & binarised & 0,88\% \\
\end{tabular}
\label{pos}
\end{center}

\caption{Performance of the classification task depending on which layers of the network are quantized, on the MNIST database. Layer 0 is the input layer, whereas layer 3 is the output one.}
\label{pos} 
\end{figure}
\vspace{-1.25cm}

	\subsection{Performance comparison}

\begin{figure}[!h]
		  \begin{center}
                    \begin{tabular}{cc}
                    \includegraphics[width=6cm]{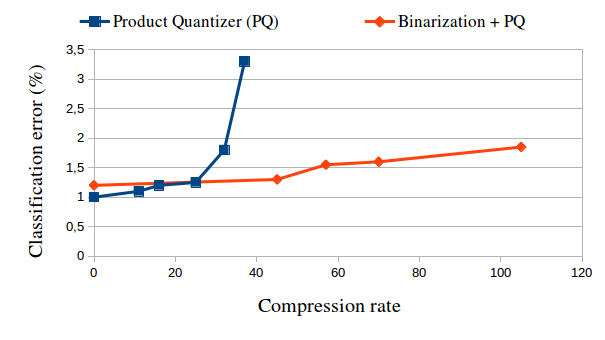}&
                    \includegraphics[width=6cm]{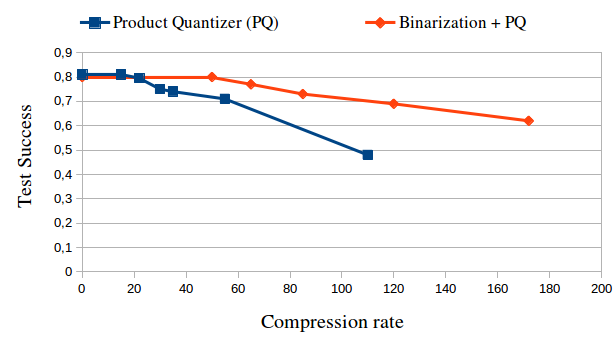}
                    \end{tabular}
		    \caption{Comparison of our proposed method with previous work on the MNIST dataset.}
                  \end{center}
                  \label{overall}
		\end{figure}

		Our second experiment shows a comparison with previous work. The results are depicted in Figure \ref{overall}. Note that in both cases compared networks have the exact same architecture.

		As far as our proposed method is concerned, we choose to compress only the two outputs layers, which are fully connected. Since their sizes are distinct, we are not able to use the same PQ coefficients $k$ and $m$ twice. Note that layer 2 contains almost all weights and is therefore the one we chose to investigate the role of each parameters.

		We observe that our added regularization cost allows to significantly improve performance. For example for the MNIST database, if we want to respect a loss of 2\%, we have a compression rate of 33 with single PQ, whereas our learn-based method leads to a compression rate of 107. 

		This compression rate concern only the two output layers. Howewer, as the output layers contains almost all weights, we still have a significant compression: on this specific example, using our proposed method the memory used to store the network weights fall down from 26MB to 550kb.

\section{Conclusion and Perspectives}

In this paper we introduced a new method to compress convolutional neural networks. This method consists in adding an extra term to the cost function that forces weights to become almost binary. In order to compress even more the network, we then apply Product Quantization and the combination of both allows us to reach performance above state-of-the-art methods.

We also demonstrate the influence of the depth of the binarized layer on performance. These findings are of particular interest to us, and a motivation to further explore the connections between actual biological data and deep neural systems.

In future work, we consider applying this method to larger datasets (e.g. ImageNet). Such datasets typically require larger networks, leading to an increased interest in obtaining good compression factors. In addition, these network are expected to be deeper, and thus allow studying thoroughly the impact of binarization depending on the deepness of layers. We also consider exploring more complex regularization functions, in particular in order to extend our work to $q$-ary values, $q$ being layer-dependent and determined on the fly.

Finally, the next step consists in making activities of neurons also binary. With both connections and activities binary, one could propose optimized digital implementations of these networks leading to higher throughput, lesser energy consumption and lesser memory usage than conventional implementations.

\subsubsection*{Acknowledgments}
This work was founded in part by the European Research Council under the European Union's Seventh Framework Program ( FP7 / 2007 - 2013 ) / ERC grant agreement number 290901.

\bibliography{header}
\bibliographystyle{IEEEtran}

\end{document}